# Before we can find a model, we must forget about perfection

Dimiter Dobrev
Institute of Mathematics and Informatics
Bulgarian Academy of Sciences
*d@dobrev.com*

With Reinforcement Learning we assume that a model of the world does exist. We assume furthermore that the model in question is perfect (i.e. it describes the world completely and unambiguously). This article will demonstrate that it does not make sense to search for the perfect model because this model is too complicated and practically impossible to find. We will show that we should abandon the pursuit of perfection and pursue Event-Driven (ED) models instead. These models are generalization of Markov Decision Process (MDP) models. This generalization is essential because nothing can be found without it. Rather than a single MDP, we will aim to find a raft of neat simple ED models each one describing a simple dependency or property. In other words, we will replace the search for a singular and complex perfect model with a search for a large number of simple models.

## Introduction

When exploring an unknown city, we try to chart a map of that city. Similarly, to understand an unknown world, we try to construct a model of that world. The model is similar to a map of the world and takes the form of a directed graph.

What model of the world are we looking for? Is it a generator or a descriptor?

A generator means that the model we are looking for must be perfect and must provide a complete description of the world. We need a full description of the world in order to create (generate) the world. Well, we do not need to create the world because it is already created so all we need is to understand it.

We will try to find a descriptor which describes the world partially by saying something without saying everything. If the world is complex enough, it cannot be fully understood. Therefore, we will abandon the search for a generator (full description of the world) and will focus on finding a descriptor (some partial description).

In this article we will assume that there is a generator model which describes the world completely. We will use the generator for a theoretical purpose so that we can define what is an *event*. Although we will use the generator, we will not attempt to find it because we assume that the generator is far too complex to be practically found. Instead of looking for such a model, we will try to find a handful of simpler models, each one describing a specific dependency or property. We shall call them Event-Driven (ED) models.

Similar to most other authors, we will assume that the generator model of the world is Markov Decision Process (MDP). As regards ED models, they are a generalization of the MDP where instead of actions we have events.

To the best of our knowledge, the first attempts to introduce Event-Driven models was made in articles [1] and [2]. However, the approach in our article differs from the one in articles [1, 2], because their authors use events in order to find a policy, while we use events in order to find a model.

From the MDP we will remove something which we do not need for the time being – the goal. Therefore, we will remove the rewards and discount factor. The outcome will be Reinforcement Learning without reinforcement. Why should we dispense of the goal? Because in this article we aim to find a model of the world rather than a policy. That is, the question we ask is "What is going on here?" instead of "What should I do?". The goal is indispensable for answering the second question, but we do not need it for answering the first one.

Event-driven models will be defined as a generalization of MDP models. To this end, we will start with the simplest model and will generalize it in several steps by going through MDP models until we arrive at ED models.

The simplest possible model we can start from will be the Fully Observable Markov Model (FOMM). This is the simplest version of Markov chain. We will see how the FOMM can predict the past. We will also see that a standard FOMM can be found for any world, although this model will not be perfect in the general case.

Our next step will be to demonstrate that the agent needs dynamic memory, while a Fully observable model means a memoryless model. This will be our reason to forget about Fully observable models. We will make our first generalization and will then move on to a Partially Observable Markov Model, known in literature as Hidden Markov Model.

We will provide a formal definition of the concepts *fact* and *event*. These will be defined as a subset of states and a subset of the arrows of the generator model. We will demonstrate that these concepts are similar, although somewhat different.

Is there just one unique generator? We will demonstrate that this is not case and that there can be a minimum and a maximum models. By minimum and maximum we do not mean the number of states but the "knowledge" which these states have.

Which generator shall we use to define *facts* and *events*? The answer is "Some of the generators". One should however to be aware that for each event a generator which presents that event must be sought. (An event can be a subset of the arrows of one generator, but not a subset of the arrows of another generator.)

Our next generalization will be the MDP model. This will be a Partially Observable Markov Decision Process, from which we will remove the rewards and discount factor. The new feature of this model will be that it takes into account the agent's actions. From that model we will remove the constraint that the event is only one (in other words, the requirement that all arrows are same).

We will note that the agent in the MDP has free will (i.e. the agent can do whatever he likes), while the world is limited by certain policy.

On this basis we will construct two versions of the MDP. In the first version, both the world and the agent are deprived of free will and are bound by certain policies. (We will name that model MDP Fixed). In the second version, both of them enjoy free will and can take any action they wish. That is, they will be free to choose any of the possible moves.

Next, we will make a generalization of these three models (MDP and its two versions). We will call that generalization MDP Plus and will demonstrate that it is a quasi-perfect model. In other words, we will abandon perfection to a certain extent, but not all of it just yet.

We will introduce the concept of *preference*. The model tells us what can happen, while the preference indicates what the agent prefers to happen. The preference is the lever by which the agent will influence the world.

We will develop an inverse MDP Plus model which predicts the past and will see that the inverse MDP model is not an MDP model – which is the reason why other authors do not consider inverse models or predictions of the past.

After that introduction and consideration of perfect and quasi-perfect models, we will completely abandon perfection by discarding the Markov property.

The next step will be to replace the agent's actions with some events. This will be the most important generalization from which we will derive the Event-Driven models.

For the model to make sense, something special should occur in its states which distinguishes one state from another. Will name this special occurrence a *trace*. We will demonstrate that the trace can be imperfect, too.

In our terminology, a dependency which occurs from time to time will be a *phenomenon*. We will introduce a trace which has a memory. For example, a state can memorize that some object has come to inhabit that state. It can also memorize the phenomenon observed by the agent when he was in that state the previous time and display the same phenomenon to the agent the next time he comes to that state.

We will introduce the object as an abstraction. The object will have certain properties, which we will present as Event-Driven models.

We will look at the relation between the generator and the Event-Driven model. The relation is that the ED model is the quotient set of one of the generators. The upside of quotient sets is that they significantly reduce the number of states. The downside is that much of the information about the world is lost, because the states are equalized with respect to some equivalence relation which can be very coarse.

Thus, theoretically we can obtain an Event-Driven model from the generator, but instead of going down this road we will construct the ED model directly from real events.

How shall we define these events? This does not require a formal definition but a straightforward answer to the question "When does an event really occur?" The events will be detected directly and indirectly.

The direct detection approach will rely on characteristic function, while indirect detection will rely on the trace (i.e. from what has happened we will infer that we have moved to another state and thereby that an event has occurred).

## Overview of existing publications

The earliest study about Event-Driven models we have been able to find in published literature are two articles, [1] and [2]. Here we will discuss [2], because the ideas in that article are more clear and straightforward.

In [2] the authors noted that the generator model has far too many states. It follows that a model with a lesser number of states should be found. Regretfully, the authors of [2] did not take that path, but skipped the modeling step and rushed to find a policy.

Nevertheless, [2] does what is most important. It demonstrates that the basis for understanding the world should be events rather than the agent's actions. The authors of [2] have introduced the concept of *event*. The event definition used in the present article is borrowed from the definition provided in [2] (with some clarification).

There is an important difference between the approach used in [2] and our approach. The difference is that we aim to find a model, while the authors of [2] aim to find a policy.

If you find yourself in an unknown situation, which is the first question you ask? Is it "What is going on here?" or "What should I do?". The important question is the latter. It is crucial to decide what to do, but the first question we should ask is "What is going on here?" Once we know what is going on, it will be easier to find out what to do.

In this article we aim to find model, i.e. we try to answer the question "What is going on?". The authors of [2] are looking for a policy. That is, they spearhead to the "What should I do?" question without knowing the answer to the first question.

Thus, the authors of [2] have spotted the problem and have made the first step to solving it, but have not made the second step, which is the introduction of Event-Driven models.

Many other articles dedicated to AI also search imperfect models, but typically they employ a different understanding of imperfection. These publications look for a complete (albeit imprecise) description of the world, while imperfection in this article means that the description *per se* is incomplete (partial). In the first case the model is not precise enough and sometimes issues a wrong prediction, while in the second case the model sometimes does not issue any prediction at all. Let us illustrate this difference with two college students. The first student answers all questions, even though some of his answers are wrong. The other student does not manage to answer all the questions. Both students are imperfect, but each one in his own way.

Examples of studies which propose imperfect models are those which look for a perfect solution within a space of possible solutions. When the space is excessively large and cannot be traversed from end to end, these studies propose to approximate the solution and find another solution which is close, but somewhat imprecise. In other words, it is proposed to find a solution which makes mistakes from time to time. These approximated solutions are typically derived by starting with one solution and subjecting it to multiple improvements. The efforts to find the first approximate solution and make subsequent improvements often rely on heuristics.

Another approach to the construction of imperfect models is to find a function on the basis of training cases. When the cases are too many, the resulting function is overly complex. Then, part of the training cases are discarded and a function which covers most but not all the cases is sought. This again results in a model which makes mistakes from time to time.

These two types of studies are focused on finding models which answer all questions, although some answers may be wrong. Conversely, the model in this article is imperfect because sometimes it dares say "I do not know".

## Reinforcement Learning without reinforcement

So we said that we will only try to answer the question "What is going on?" and will not look at the second question, namely "What should I do?". This means that we do not need a goal. Most authors define an MDP specifically for Reinforcement Learning by adding a goal. Therefore, they introduce a rewards and discount factor.

When aiming to find a policy we cannot do without a goal, but if the aim is to find a model only, then no goal is needed. That is, we will explore Reinforcement Learning without reinforcement. We will not have any goal as our sole goal will be to have the knowledge. The goal will not be important to us because if we come to a sufficiently good understanding of the world, then any goal will be within our easy reach.

In [6] we already discussed why the discount factor should be removed from the MDP definition. If we remove the rewards as well, we will lose the goal, but may well lose part of the information. To avoid losing information, we will move the rewards to the observation. For example, the goal of schoolchildren are the scores they earn at school. If we remove the scores, we will hide part of the information, so we can let the scores stay and tell the kids that scores are only points of reference and not a goal.

The only problem will then be that with some models the observation is labeled to the state of the world, while the reward is labeled to the transition between states (the arrow). For this reason, we will assume that we have a trace (i.e., something special occurs) not only in the states, but also in the transitions between states.

## What is given?

We have an agent and a world which interact with one another. We can describe the interaction as *observation-action* or as *question-answer* pairs.

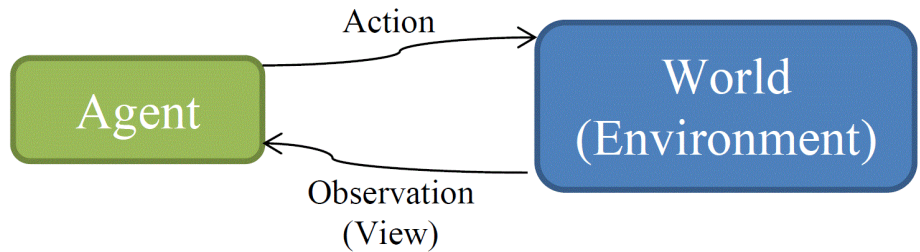

**Figure 1**

The result from this interaction is the following sequence:

$$\ldots, v_{-2}, a_{-2}, v_{-1}, a_{-1}, v_0, a_0, v_1, a_1, v_2, a_2, \ldots$$

Let the set of possible observations be $\Omega$ and the set of possible actions be $\Sigma$. Let $\Omega$ and $\Sigma$ be finite sets. The *observation-action* sequence can be thought of as a word (finite or infinite) over the alphabets $\Omega$ and $\Sigma$.

The *zero* moment here will be the current moment. Everything before that moment will be the past and everything after – the future. Most authors suggest that there is some absolute beginning: an initial moment before which no past whatever exists. We will not assume that such an absolute beginning exists. As we explained in [6], even if such an absolute beginning existed, it occurred in the far too distant past, so we had better not bother about it and deal with the present moment only. Most authors use the term *initial moment* to describe the absolute beginning, however *our initial moment* will be the *current moment*.

## What are we looking for?
We are looking for a model of the world. The form of that model will be a directed graph (Figure 2). The vertices in the graph are our states. One of these states we will call *initial* or *current*. In Figure 2, this is state 2 which is presented with a larger circle.

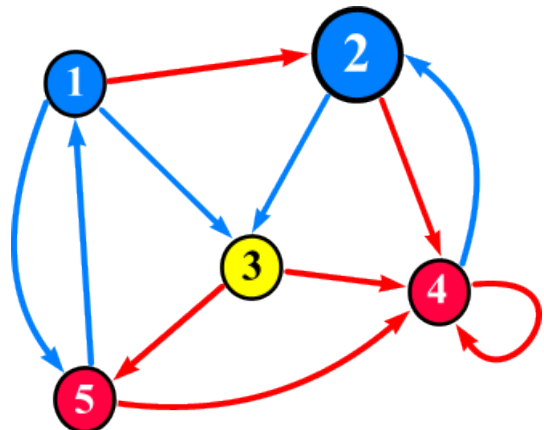

**Figure 2**

The arrows will be labeled with certain reasons which change the states of the world. In our terminology these reasons will be called *events*. In Figure 2 these events are represented by arrows of different colour.

Each arrow will be associated with two probabilities. The first one is the probability of an arrow with such colour being chosen (the probability that the event will happen), while the second one is the probability of exactly that arrow being chosen (among several arrows of the same colour). The product of these two probabilities is the probability of that arrow being actually used.

Quite often one of these probabilities will be obvious and then we will assign to the arrows only one probability. We have divided the probability in two, because choosing an arrow is a twofold process. An event must be chosen first and only then the exact arrow can be chosen.

So far we described the model, but still it does not tell us anything about the world. For a model to be meaningful, something special should occur in its states. The occurrences we expect to happen in its states will be called *trace of the model*. The trace in Figure 2 is visualized by using different colours for the various states. Thus, having two states in the same colour means that we expect the same occurrence to happen in these two states.

The purpose of the model is to tell us something about the future and the past. A perfect model would be the one which provides a perfect description of the future in case that the agent's policy is fixed. (The future is determined both by the world and the agent. Therefore, we cannot know what will happen if we do not know what the agent will do. So we assume that the agent's policy is fixed. What we want is a perfect description only of the future, but not of the past too, because the model is able to describe perfectly the future without describing perfectly the past).

Now is the time to say what is a perfect description of the future and what is a policy.

## Definitions

The *action-observation* sequence will not be uniquely defined even if the model is given. In other words, the model will give many possibilities for both the past and the future.

In our terminology, the *action-observation* sequence before the initial moment will be *possible past* while the sequence from the initial moment onward will be *possible future*. To avoid dealing with infinite words, we will consider the possible developments of the past and of the future.

**Definition.** A *possible development of the past* is any finite word which is end of a possible past.

**Definition.** A *possible development of the future* is any finite word which is beginning of a possible future.

Now we will define what is a *perfect description of the future*, in which each possible future development occurs with a precisely defined probability.

**Definition.** A perfect description of the future is the set *Future* each member of which has the form $<\omega, p>$, where $\omega$ runs the possible future developments and $p$ is the probability for the $\omega$ development to happen ($p>0$).

*Future={<$\omega$, p>| $\omega$ is a possible future development, p is the probability that $\omega$ will happen}*

The perfect description of the future tells us what is going to happen with a very high degree of accuracy. More precisely, it tells the future with an accuracy equal to some randomness (dice). Such a description can be depicted as an infinite tree with vertices associated with actions and observations, and edges associated with a probability (a precisely defined probability). It is easy to build a computer program which generates this future. The program will contains a description of the tree and an operator *random(p)* which returns 0 or 1 with a probability of $p$. Of course this can only happen when the infinite tree is computable, otherwise the program cannot include a description of the tree.

An example of a future which can be perfectly described is when we flip a coin infinite number of times. Each time the coin is flipped it will fall heads or tails with a precisely defined probability. It is not difficult to build a computer program which generates the so-described future.

**Definition.** The agent's policy is a function which, for each state and possible action, determines the probability of that action being done by the agent.

$$Policy : S \times \Sigma \to \mathbb{R}$$

**Definition.** A deterministic policy is a policy which provides a probability of 1 for one of the possibilities and a probability of zero for all other possibilities.

When referring to a policy, the majority of authors actually mean a deterministic policy. If we flip a coin to determine whether we should turn right or turn left, this is a policy but not a deterministic policy.

**Definition.** A perfect model is the one which provides a perfect description of the future in case that the agent's policy is fixed.

# Fully Observable Markov Model
This is the simplest possible model of the world.

**Definition:** A Fully Observable Markov Model (FOMM) is:
$S = \Omega$ (the set of states coincides with the set of possible observations)
$p : S \times S \to \mathbb{R}$ (the probability of a transition from one state to another )

$p(i, j) = Pr(V_n=j \mid V_{n-1}=i)$ (the probability of a transition from state $i$ to state $j$) Here $V_n$ is the sequence of observations.

In literature this model is referred to a Discrete-Time and Time-Homogeneous Markov Chain. Discrete-Time means that the process is stepwise and Time-Homogeneous means that the probabilities do not depend on the step at which the process is. Figure 3 illustrates such kind of Markov Chain.

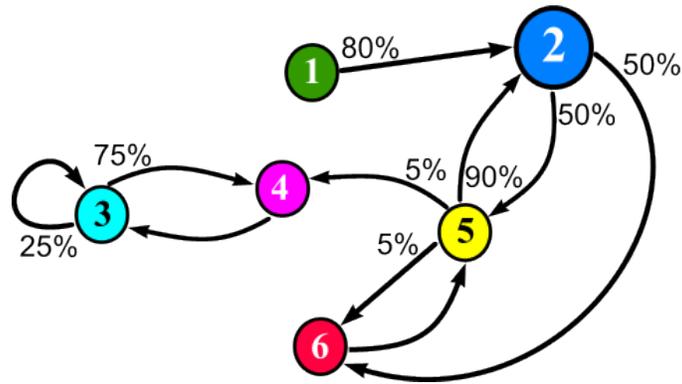

Figure 3

The arrows in this case have only one colour because there is only one event. This is the event "true" (this event occurs always, i.e. at every step).

The trace of the model in this case will be total. That is, something special will occur at each state. The occurrence in the state at hand is that the agent will see a precisely defined observation (one of the members of $\Omega$). We will assume that we have Full Observability, i.e. that we can infer the state of the model from the observation or, in other words, there cannot be two states with the same colour (i.e., with the same observation).

Each arrow will be associated with a probability. (There will not be two probabilities as the probability for that event to happen is 1 simply because there is only one possible event. This means that we are left only with the second probability. In Figure 3 the probability is indicated only if there is more than one arrow, because a single arrow means a probability of one.)

Is the Fully Observable Markov Model (FOMM) a perfect model of the world? Yes, but only if it satisfies with the Markov property.

If the future depended only on the state of the world we are in, but not on how we have arrived at that state, then the model would be perfect. The Markov property means that the model cannot be improved. If a state can be reached through two different paths and these two paths trigger different future developments, then we could divide the state in two – one state reachable through one of the paths and the other state reachable through the other path. Thus we could improve the model because the two new states would help make a better prediction of the future. If we had the Markov property, the two new states would issue the same prediction of the future, the bottom-line being that we cannot improve the model in this way.

## FOMM Inversed

So far the FOMM can issue predictions of the future. Now, can we inverse the FOMM so that we can predict the past? If we move against the arrows, we will see the possible past developments, but we wish to know more than that – not only which developments are possible and which are impossible, but also the probability of each possible development.

For each state and for each outbound arrow we have a probability of exiting that state through that arrow. From these probabilities, can we find what the incoming probabilities are? Namely, for each state and for each inbound arrow, can we establish the probability for us to have entered that state from that arrow? The answer is yes, provided that there are no "white peaks". But before that we should define what a *white peak* is.

**Definition.** A *black hole* is a set of states wherein no path leads from a state inside that set to a state outside that set. Another requirement of this definition is that the set is not empty and does not contain the initial state.

In other words, once we enter this set of states, we stay there forever because there is no way out of that set. A *white peak* is the opposite of a *black hole*. Once we exit a *white peak* set, we shall never return to that set again.

**Definition.** A *white peak* is a set of states wherein no path leads from a state outside that set to a state inside that set. Again, that set must not be empty and must not contain the initial state.

In Figure 3 we can see both a white peak (set {1}) and a black hole (set {3, 4}).

We will assume that in FOMM there can be white peaks and black holes, but can we have sets that are *both* white peaks *and* black holes? In other words, can there be encapsulated states which one cannot neither enter nor exit? We will assume that there are no such states because they are redundant. The white peak will not contribute to predicting the future and the black hole will not be used for predicting the past. Where a state is *both* a white peak *and* a black hole, it will not be involved in predicting either the past or the future. Therefore we shall assume that these redundant states have been removed and are simply not there. The only issue with removing these redundant states is a likely breach of the rule that the sum total of the probabilities of the arrows must be one. This can happen only inside a white peak. This is exactly what happens on Figure 3 – from state 1 there is one arrow which exits the state with a probability of 80%. Supposedly, the remaining 20% go to redundant states which we have already removed. However, this breach would not be a problem, because outbound probabilities in white peaks are irrelevant since they are used for predicting the future, but white peaks are not involved in predicting the future. Similarly, these considerations can be applied to black holes and inbound probabilities, which we are going to introduce right now.

**Theorem 1.** For each FOMM which does not contain white peaks we can calculate the inbound probabilities from the outbound probabilities. In this way we can predict the past. If the prediction of the future has been perfect, the prediction of the past will again be perfect.

**Proof:** The idea is simply to reverse the arrows and thus obtain a new FOMM which will be as good in predicting the past as the original FOMM has been in predicting the future. This will require us to recalculate the probabilities and replace the outbound probabilities with inbound ones.

The approach we will use for this proof will be that of engineers rather than of mathematicians, so we will work with numbers instead of probabilities.

In our assumption, we start from the initial state and walk down the arrows until we return to the initial state or end up in a black hole. Let these be multiple journeys, for example 100. We will count the number of times we have walked down each arrow. From these numbers we will calculate the outbound probabilities and these will be the ones to start with (with a certain statistical error). We will also be able to calculate inbound probabilities (again with some statistical error).

If we take the outbound arrows from the initial state and count the number of times we have walked down these arrows, the final sum will be 100 because we had 100 journeys starting from the initial state. Well, will the sum total of the inbound arrows leading to the initial state be 100 again? No, it may be less because we have to subtract the journeys which end up in a black hole.

Thus we can calculate the inbound probability of all states which are not part of a white peak or of a black hole. We assumed that this FOMM does not have white peaks, while black holes are not involved in predicting the past and their inbound probabilities are not important. In other words, the so-derived new FOMM will be able to predict the past.

If the original FOMM is perfect and cannot be improved, then the so-derived new FOMM will also be perfect and not liable to improvement. This statement may be intuitively true, but in fact it needs rigorous proof.

So far we saw that we can obtain a perfect description of the past from a perfect description of the future. Yes, this is true, provided however that there are no white peaks. But what if there ARE white peaks? Then the statement is no longer true because over the white peaks we can apply whatever inbound probabilities we wish.

Let us repeat the proof of Theorem 1 but now with the assumption that there are white peaks, too. We started 100 journeys from the initial state. Let us now take an arrow which comes from a white peak. We can start as many journey from that arrow as we wish – zero, one, one hundred or one thousand journeys. If we do not start any journey from that arrow, this will mean an inbound probability of zero for that arrow. The more journeys we start from the arrow, the higher probability we get for that arrow. We will do that for all arrows coming from white peaks. We will start and continue until we go to the initial state or end up in a black hole. This will change the inbound probabilities not only of the states to which we go directly from a white peak, but of all states we have been through.

Therefore, we can assign a random inbound probability to the arrows coming from a white peak, which in turn will change all inbound probabilities in the inverse FOMM. The inbound probabilities inside the white peaks are not determined in this way, but we can assign random values to these probabilities (which will not change the remaining probabilities).

In sum, we saw that, in the general case, having a perfect prediction of the future does not mean we can derive from it a perfect prediction of the past. A perfect prediction of the past we can derive only in the special case where no white peaks are present.

## The standard FOMM

A FOMM model can be constructed for each and every world. To do so, we will take the set of states Ω and will assign probabilities to the arrows on the basis of statistical data over a certain period of time. Collecting statistical data is not a problem in this case because we enjoy Full Observability, meaning that at all times we know which state we are in. A problem would arise if we gather statistics in one period and try to use it in a model which describes another period. The problem is that these two periods can generate different statistics. We will assume that the two periods coincide. Thus, our model will describe the period of time in which the statistics were collected. This will ensure that our statistics are adequate.

The question actually is how adequate will this model be? This will be a model that tells us something about the world. It will tell us the probability of any two observations occurring one after the other. That will be an average probability, but will the model be a perfect one? That is, can it be improved?

Yes, in a very special case the model will be perfect and not liable to further improvement, but in the general case there will be better models, too. Hence, the so-derived model may have the Markov property, but this is highly unlikely. For the model to be perfect, the statistical data should cover an infinite period of time because each finite sequence has a model which describes it completely, meaning that its model can be improved.

Even infinite intervals can be described completely (albeit by infinite models). For a model to be non-improvable any further, we must have a set of infinite intervals with the cardinality of the continuum. Only then the Markov property will be applicable.

Let us have a sequence of *black* and *white* where the next observation is always determined by flipping a coin. Then, if we construct the standard FOMM, it will return *black* with a probability of 50% and that will be the perfect model. Imagine now the world gives us two times *black*, then two times *white* and so forth. Then the standard FOMM will be the same, because the probability of *black* after *white* will again be 50%, but this time it will not be a perfect model since it can be improved so that it can predict more accurately what we will see next (indeed with an accuracy of 100% in this case).

The truth is that to improve the model we need memory which in turn requires us to abandon Full Observability.

## The dynamic memory

What is the difference between a fixed and a dynamic memory? There are seven days in the week. This is a hard fact which you can memorize once and remember it forever by storing it in your fixed memory. On the other side, today is Thursday. You cannot memorize it forever, because tomorrow it will be Friday and Thursday will no longer be true. The fact that today is Thursday you should store in your dynamic memory which you should change on a regular basis.

Do we need a dynamic memory? If we lived in a fixed world where it is always Thursday morning, we have always had breakfast, stay at the same place at all times and nothing else changes, then we do not need a dynamic memory. But in a changing world we need a dynamic memory if we are to figure out what is going on.

We will store the model in the fixed memory and will use the dynamic memory in order to memorize the current (initial) state we are in. In a certain sense, a proper memory is only the dynamic memory. Artificial Intelligence (AI) will need to discern the model and store it in its fixed memory. But if we construct a device which is not AI and is not apt for every world but is custom-designed for a particular world, the device may have the description of the model embedded in itself and will need to memorize only the state it is currently in. Therefore, a device which is custom-built for a specific world can do without a fixed memory as it needs only a dynamic memory.

Therefore, the fixed memory can be part of the hardware. Examples include computers with RAM (read and write memory) and ROM (read-only memory). The fixed memory is ROM and the dynamic memory is RAM. In a certain sense, a genuine memory is only RAM because ROM is part of the hardware rather than a genuine memory.

How big is the model's dynamic memory? How many bits? The answer is a binary logarithm of the number of states in the model. More precisely, not the number of states, but of the maximum number of states with the same colour, because we do not need to remember what we observe right now (the colour of the state), but only which of the several possible states we are now in.

How big is the dynamic memory with Full Observability models? The maximum number of same-colour states in these models is one. The memory size, therefore, is zero bits.

We are convinced that the interesting worlds are not fixed (they change their condition). Accordingly, we will need models with memory. For this reason we will abandon Full Observability and will only consider Partial Observability models. This distinguishes us from most other authors who prefer to deal with Fully observable models as they assume that these models are more simple. We believe that the general is simpler than the special so Partialy observable models are more simple, more understandable and more functional.

Recently, AI developers have made major progress in the area of recognition, but their applications still have no clue about what is going on. For example, many AI applications are great at recognizing faces or voices, but fail to maintain a basic conversation.

The reason is that most researchers use memoryless models (neuron networks and Fully observable models). To recognize faces or voices, one does not need a dynamic memory. If you see the same face twice, you are expected to say the same name. However, if you wish to have a basic conversation, you do need a dynamic memory because if they ask you the same question twice you are not supposed to give the same answer. You should remember that the question has already been asked.

That is why we abandon Full Observability and move on to the next models.

## Hidden Markov Models

The only difference between Hidden Markov Models (HMMs) and FOMMs is that with HMMs two or more states may have the same colour (these are states in which we see the same observation).

**Definition:** A Hidden Markov Model is:
*S* (set of states)
*Trace : S → Ω* (what the agent sees in each state)
*p : S × S → ℝ* (the probability of a transition from one state to another)

$p(i, j) = Pr(s_n=j \mid s_{n-1}=i)$ (the probability of a transition from state *i* to state *j*)

Here $s_n$ is the state at step $n$.

Unlike the FOMM, the Hidden Markov Model has memory. This takes us to the next question: What exactly do HMM states *memorize*? In this article, however, we will discuss what do these states *know* about the past and the future. We will not use the verb *memorize* because it implies that there is only one single past which we have memorized, i.e. certain occurrences have happened and we have stored them somewhere. In our case, instead of memorizing the past, the model will help us say a few things about the past and the future. Most notably, we will be able to say which developments are possible and which are not. Perhaps we will know the probabilities of some of the possible developments.

What can a state know about the past and the future? It may know certain facts. In other words, if the state is the current one, what can we say about the past and the future? What we can say are certain facts.

But let us first say what is a fact and what is an event.

## Facts vs. Events

A fact is something which can be either true or false. An event is something which sometimes occurs and sometimes does not. In [6] we defined an event as a Boolean function of time. The same definition can be applied to facts and then facts and events will be the same thing from a theoretical perspective.

Nevertheless, in our understanding an event is something which is true from time to time while facts are true in certain intervals of time. This is an informal idea of having two different objects, which may not be different if treated as Boolean functions.

In this article we will modify the definition of an *event*. We will borrow the definition provided in [2] and will define an event as a set of arrows. A fact we will define as a set of states. Thus, facts and events will be formally different, but still very similar objects.

**Statement 1.** Facts and events are largely similar.

**Justification:**
We will demonstrate that any fact can be expressed as an event. Let us take the set of outbound arrows from the states of that fact. This will be an event which will occur at the moments when the fact is true. (Had we taken the inbound arrows which enter the states, we would have had the same event, but shifted one step backward.)

Similarly, we can express the event $E$ as a fact, however this will not be a fact in the same model, but in another model which is equivalent to the first one. We will construct the new model by doubling the states of the first model. We will replace each state $s_i$ with $s_i'$ and $s_i''$. Each arrow $s_i \rightarrow s_j$ will be replaced with two arrows:

$s_i' \rightarrow s_j''$ and $s_i'' \rightarrow s_j''$ if $(s_i \rightarrow s_j) \in E$
$s_i' \rightarrow s_j'$ and $s_i'' \rightarrow s_j'$ if $(s_i \rightarrow s_j) \notin E$

The fact will be the set $S''$, where $S'' = \{s \mid \exists i : s = s_i''\}$.

This construction is not very good because the fact will be true not at the step at which the event occurred, but at the next step. Moreover, it is not quite clear which the new initial state should be. If the initial state was $s_0$, which should be the new state – $s_0'$ or $s_0''$? We will however shut our eyes to those structural imperfections.

∎

Why have we changed the definition of an event? Imagine you have a statuette and somewhere on this statuette there is crack which you wish to describe. Imagine now that you do not have the full statuette, but only a slice of it. In this case you will be able to describe the crack only on the basis of that slice, which will be an incomplete and inaccurate description. It may be that no part of the crack is present in your slice, but this does not mean the crack is not there. Anyway, if all you have is the slice, that will be all you know about the statuette. You will not be able to see the remaining part of it, but still you can picture it in your mind. Therefore the crack can be better described as a three-dimensional object which forms part of an imaginary statuette.

Similar is the situation with the world and your life. If we have only one life and decide to describe the event on the basis of that concrete life, the event would have the form of a Boolean function. If the event does not happen during our life, it would not mean that the event is impossible to happen in that world. Had we lived our life differently, the event might have happened.

Accordingly, we will define the event as a set of arrows in the generator model. (Earlier we assumed that a generator which provides a perfect description of the world does exist.)

The problem is that it is not a unique model. There are many other equivalent models.

## Minimal models

**Definition.** Two models of the world are equivalent if they tell us the same story about the past and the future of the initial state.

We do not mean trivial equivalence where the models are isomorphic. We will consider models with more knowledgeable states as well as models with less knowledgeable states.

A minimal model is a model with states the knowledge of which is limited to the minimum. With these models, two states tell us the same story about the past *iff* they tell us the same story about the future. (If two states know the same things about the past and the future, these states are equivalent.)

As regards the past, the states in the minimal model will not "remember" anything which is redundant. "Redundant" is any fact which cannot have any impact on the future. It is self-evident that it does not make sense to memorize a fact if the future does not depend on that fact. (So, if two states tell us the same story about the future, they will tell us the same story about the past because the redundant facts are not "remembered").

Similarly, if we apply the same logic to the future, the states will not know anything about the future which does not follow from the past. (In other words, the state cannot know a thing if it has nowhere to learn it from.) This means that if two states tell us the same story about the past, they will tell us the same story about future. (If their possible developments in the past coincide, their possible developments in the future also coincide.)

What can disrupt the minimalism of the model? What can cause a state to know more than the minimum? There are two possible causes:

The first one is nondeterminism. If we start from the initial state, a few steps later we will arrive at a set of possible states. (If the set includes probabilities of the individual states, this is called "belief".) If the set has states with different future, then these states know more than the minimum, because the "belief" tells us that we are in one of these states, but we still do not know which state exactly we are in. If we know the exact state we are in, then we know about the future something we have nowhere to learn it from.

Second, there may be a situation where two states have different pasts but an identical future. Then, if we know which of these two states we are in, we would know something redundant about the past – something, which the future does not depend on.

If we make the model deterministic we will eliminate the first cause and if we minimize the model will get rid of the second one. Therefore, the minimal model is determinized and minimized in either direction of the arrows (downstream and upstream).

In [5] we discussed the minimal models in detail. In [5] we also described an algorithm for finding a minimal model such that whichever state we start from (downstream or upstream) we get minimalism and determinism. Regretfully, the algorithm in [5] is wrong. One can see that the elimination of nondeterminism in a forward direction induces nondeterminism in a backward direction and vice versa, which is why the algorithm in [5] does not work.

Generally, a minimal model in which we can start from any state does not exist. Nevertheless, there is a minimal model where, if we start from the current state we will have minimalism and determinism all the way forward. Similarly, we will have the same if we go backward.
This model can be constructed in three steps:
    1. Construct a model which is minimal and deterministic all the way forward from the initial state. (In literature the deterministic model is known as Belief MDP. So we construct this model and minimize it by merging the states that have the same future. In other words, we merge the states whose possible developments of the future coincide and these developments are equally probable.)
    2. Construct the inverse model which will be unique unless there are white peaks (if there are white peaks, take one of the possible inverse models.) Determinize and minimize that model (as in step 1). Take the resulting model and build from it an inverse model (this model will be deterministic and minimal in a backward direction).
    3. Finally, assemble the two models in one. To this end, introduce new initial states in the two models and join the models at the new states.

The three elements in Figure 4 illustrate these three steps:

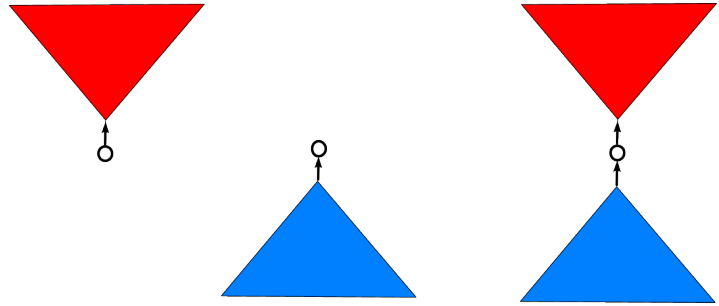

Figure 4

Thus, the future will always be determined only by the first model and the past will always be determined only by the second model. This is due to the introduction of new initial states. This guarantees that once we exit the initial state we can never return to it. (Hence the first model is a black hole and the second model is a white peak.)

The so-constructed model is minimal in a sense that the states know as little as possible, but not in a sense that the number of states is the least possible. If we are to construct a minimal model with a minimal number of states, we need to combine the two models and use some of the states for predicting both the past and future.

In [5] we discussed a minimal model case where the arrows have no probabilities. The good thing about this case is that from a finite model we obtain again a finite model. If we were to minimize the MDP, then from a finite model we may end up with an infinite model. The reason is that the subsets of a finite set are finitely many, but the "beliefs" are infinitely many. (In other words, the minimization will not reduce the number of states, indeed it may even produce a considerably larger number of states.)

## The selection of the generator

We assumed that a generator which provides a perfect description of the future does exist. We saw that the generator is not unique. In [5] we saw that there is a minimal model the states of which do not know anything redundant, and a maximal model the states of which know everything about the past and the future.

We may assume that there are models the states of which know even more than the states of the maximal model. "What more than everything?", one might ask. They may know some insignificant facts which have no bearing on either the past and the future. For example, "Is there life on planet Mars?". Let us assume that the answer to this question is a fact which has no influence whatever on your past or on your future. In this case, it is an insignificant fact. We will assume that we are not interested in insignificant facts and accordingly we will not consider models with such facts.

Insignificant facts needlessly complicate the model. For example, the fact "There is life on planet Mars" can be presented by duplicating the model. This will give us two identical models and we will be in the first model if there is life on Mars and, respectively, in the second model if there is not. So the model will know the answer of this insignificant question, but nothing will follow from it because the past and the future in the two models are identical. In addition to the insignificant duplication of the model, there can be insignificant branching. For example, if life emerges on Mars, there will be an insignificant branching in our model. If we go left, we will know that life is there already, and if we go right we will know that there is no life yet. This knowledge will again be insignificant, because it has no bearing on either our past or future.

Another example for insignificant event is: "One butterfly waved with wings." There is a theory that this event can be significant and lead to a hurricane, but here we will assume that this event has no effect on our future. We'll also assume that we didn't notice when the butterfly waved with wings. That is, this event did not affect our past either. This means that the event is insignificant.

So far so good. Now, let us select the generator on the basis of which we will define facts and events. The first appropriate candidate is the minimal model, but this model does not memorize any redundant thing. Why not afford remembering redundant things? We do not know in advance what is useful or useless, so we memorize many redundant things. For example, in our world it is important to know which day it is by modulo 7, because these are the days of the week. But nobody cares about counting days by modulo 2 because it does not matter in our world. Just in case, let us not prohibit people from counting days by modulo 2 because one day this may prove useful.

Furthermore, the minimal model is not unique. While it is true to say that the minimal model with the least number of states is indeed a unique model, but the states of that model can be a lot, because the minimization can increase the number of states. (For example, a minimization exercise can cause a finite model to explode into an infinite model.) The states of the minimal model know as little as possible, but this does not mean that the number of these states is reduced to the minimum.

For these reasons we will not choose the minimal model. Shall we then choose the maximal model? Again no, because this model is far too complicated. Its initial state is "belief" which is a set with the cardinality of the continuum. One thread, which corresponds to one possible life, goes through each of the states of the "belief". If we take only one thread (only one of all the possible lives), then the event will be a Boolean function, but as we said, this does not make us happy.

Thus we decide not to choose the exact generator. Let it be a generator which resides between the minimal and the maximal model. The event will be a subset of the arrows of that generator.

This definition is good because we do not intend to use it. We need a formal definition of the term *event* and we will define it as part of the arrows of some generator. We do not specify which generator it is, but we already said that we will not try to find it anyway. So let it be any of the generators.

This is our clarification to the definition provided in [2] which says that the event is a subset of the generator's arrows, but does not specify which the exact generator is. Perhaps the authors of [2] assume that there is just one generator, but this is not the case.

## MDP Models

The deficiency of FOMMs and HMMs is that these models do not take into account the agent's actions. There may be a world where the agent's actions have no impact on what is going to happen. It may be that the agent's actions matter, but the model describes a dependency which is independent from these actions. Hence, FOMMs and HMMs may be useful, but they are not sufficient in the general case. Our aim is to find a model which can accommodate the agent's actions and accordingly we move to the next models. These are the MDP models.

Let us note that we mean "Partially observable" rather than "Fully observable" models. We will omit the adjective "Partially observable" assuming that unless "Partially observable" or "Fully observable" is explicitly mentioned, the default understanding is "Partially observable". This distinguishes us from most other authors who assume that in the absence of explicit mentioning, the default understanding is "Fully observable".

The difference between HMM and MDP models is that HMMs allow for only one possible event (this is the event "true"), while the possible events with MDPs are $\Sigma$ (the agent's actions).

Both in HMMs and MDPs we can define the trace either as a concrete observation which we must see in that state or as a set of possible observations each one having a precisely defined probability. The two definitions are equivalent which is why for the HMMs we have selected the first one because it is more simple while for the MDPs we will select the second one since it produces a model with less states. (For the minimal models we will assume that the trace gives only one observation, because if it gave several possible observations, that would mean nondeterminism, while a characteristic feature of minimal models is their determinism.)

**Definition:** An MDP model is:
*S* (the set of sets )
*Trace : S × Ω → ℝ* (the probability of the agent seeing a concrete observation in a concrete state)
*Agent : S × Σ → ℝ × ℝ* (the probability of the agent choosing a particular action or, more precisely, the interval in which this probability resides. This interval will always be [0, 1].)
*World : S × Σ × S → ℝ* (the probability of a transition from one state to another state through a certain action)

*World(i, a, j) = Pr($s_n$=j | $s_{n-1}$=i, $a_{n-1}$=a)* (the probability of a transition from state *i* to state *j* through action *a*)
Here $s_n$ is the state of step *n*, and $a_n$ is the action at step *n*.

The *Agent* function is completely needless in this definition. We do not need a function which always returns the [0, 1] interval. Why would we need a function which is a constant? We have added this function for the sake of the two versions of the MDP model which we are going to construct (MDP Fixed and SMDP), since in these versions this function will not be a constant.

Our definition does not include rewards and discount factor, but as we explained earlier we do not need it for the time being.

**Note:** Importantly, for each world there exists an MDP model which provides a complete and unambiguous description of that world. I.e. there is a perfect MDP model. There may or may not exists a finite MDP model, but an infinite MDP always exists. Accordingly, we can assume that the world has a generator and that generator is MDP. (The same applies to HMMs provided that we limit the worlds to those which are not influenced by the agent's actions.)

# Free will and constraint

The MDP model is perfect because it tells us all about the future provided that the agent's policy is known. Why the MDP does not fix the agent's policy? Because we assume that the agent is not part of the world and enjoys free will (i.e. can do whatever he wishes), while the world is constrained by some rules which define exactly the world's next moves (with an accuracy equal to a certain probability).

Thus, we will assume that the agent has free will, while the world does not. Imagine that you are required to flip a coin and drive left or right depending on which side of the coin is up (heads or tails). This means that you don't have free will because you have to obey the coin.

Let us imagine that the agent and the world are two players who play a game or two interlocutors engaged in a conversation. We will note then that the MDP model does not provide a level playing field for these two players. The agent is free to do whatever he wishes while the world is tied to a predefined policy. The agent can choose any of the possible moves with whatever probability he likes (i.e. with a probability in the interval *[0, 1]*), while each of the world's possible moves is associated with an exactly defined probability and the world is required to select that move with the so-defined probability. In other words, the agent enjoys free will while the world's freedom is limited to only one policy.

You may be deprived of free will and bound to follow a precisely determined policy. There is, however, another scenario: you may have free will which is confined within certain limits. Thus, you are free to select your policy within certain constrains.

**Definition.** A constraint on the agent is a function that for each state and possible action provides an interval within which the probability of the agent performing that action resides.

$$Constraint : S \times \Sigma \rightarrow \mathbb{R} \times \mathbb{R}$$

An absolute free will exists when the interval in question is [0, 1]. Free will is completely absent when the length of the interval is nil (i.e. when the probability is exactly defined).

The policy of the world can be defined in way similar to the agent's policy:

**Definition.** The policy of the world is a function which for each state, agent's action and new state returns the probability of the world moving on to the new state in response to that action of the agent.

$$Policy : S \times \Sigma \times S \rightarrow \mathbb{R}$$

The constraint on the world can be defined similarly to the constraint on the agent. The *Agent* and *World* functions in the MDP definition represent the constraint on the agent and respectively the policy of the world.

Can the world have free will? Imagine that another agent, who enjoys free will (unconstrained agent), resides in that world. The actions of the unconstrained agent will manifest themselves in the behaviour of the world. The unconstrained agent's freedom to do whatever he wants does not imply that the world will do whatever the world wants. The world may have a degree of freedom, but still remains constrained by a certain constraint.

As far as the world is concerned, we can replace "free will" with "unpredictable randomness". In some cases we know exactly what is going to happen. In other cases we do not know exactly what is going to happen, but know the probability with which something may happen. If even that probability is unknown to us, then we are faced with unpredictable randomness. In the previous example, when an unconstrained agent lives in the world, the next doings of that agent will be unpredictable randomness.

## MDP Versions

Now let us consider an MDP version where both the world and the agent are bound to follow a certain policy. Thus, neither of them enjoys free will. We will name this model MDP Fixed. The only difference with the MDP will be that the *Agent* function will not return the [0, 1] interval, but a concrete value (i.e. it will return the nil-length interval). MDP Fixed is a perfect model because it tells the full story of the future.

Now let us look at yet another MDP version where both the world and the agent enjoy free will. In this MDP version the *Agent* function will return the interval [0, 1] or the values 0 or 1 (when the action is impossible or respectively there is only one possible action). Similarly, the *World* function will return not a single value, but the interval [0, 1] or the values 0 or 1 (when the transition is impossible or respectively there is only one possible transition). We shall call this version State Machine Decision Process (SMDP). This name has been chosen because of its close similarity to Nondeterministic Finite-State Machine (NFSM). One difference is that in the NFSM the states are assumed to be finitely many, while the SMDP does not include such assumption. Another difference is that the NFSM deals with two state types (final and non-final), while the SMDP deals with multiple state types. (If we see a concrete observation in each state, then the state types are $\Omega$. If in each state we see several observations with a different probability, then the state types are infinitely many.) The main commonality between NFSM and SMDP is that in either case we do not assign probabilities to the arrows. The probability in the SMDP is either within the [0, 1] interval or 1 (depending on whether nondeterminism is present or absent). Therefore, the probability in the SMDP is clear and we do not need to write it down. (More precisely, the two probabilities of the arrow are clear and need not be written.)

The SMPD model is not perfect, but we can define it as quasi-perfect.

**Definition.** A quasi-perfect description of the future is the set *Future* each member of which equals <$\omega$, *[a, b]*>, where $\omega$ runs the possible developments of the future and *[a, b]* is the least possible interval within which the probability of $\omega$ to happen resides (*[a, b]* $\neq$ *[0, 0]*).

**Definition.** A quasi-perfect model is one which provides a quasi-perfect description of the future. (Here we do not constrain the agent's policy unless the model itself constrains it.)

## MDP Plus

Now let us generalize the MDP and its two versions. The full name of the resulting model will be *MDP plus free will* a*nd unpredictable randomness* (or *MDP Plus* for the sake of brevity).

In MDP Plus, both the world and the agent have free will, but it is not unlimited (i.e. it is subject to certain constraints).

**Definition:** An MDP Plus model is:
$S$ (the set of states)
$Trace : S \times \Omega \to \mathbb{R}$ (the probability of the agent seeing a concrete observation in a concrete state)
$Agent : S \times \Sigma \to \mathbb{R} \times \mathbb{R}$ (the probability of the agent choosing a particular action or, more precisely, the interval in which that probability resides)
$World : S \times \Sigma \times S \to \mathbb{R} \times \mathbb{R}$ (the probability of a transition from one state to another through a certain action or, more precisely, the interval in which that probability resides)

The difference between MDP and MDP Plus is that in MDP each arrow is associated with one probability, while in MDP Plus each arrow is associated with two intervals. In MDP we do not write down the first interval, because it is always *[0, 1]*. The second interval in MDP is nil-long so we write down a single number.

The MDP Plus model is quasi-perfect. It is a generalization of the MDP. Perfection in the MDP Plus is disrupted because we let the world have some freedom: the world is not bound to follow a particular policy and can choose its policy within a certain constraint.

## Preference

When most authors refer to a *policy* they mean that the agent is free to select any policy he wishes. In this article we refer to two policies – one of the agent and one of the world. It is legitimate to assume that the agent can chose the action he will do next, but it is not legitimate to assume that the agent can choose the behaviour of the world. On the other hand, it is not legitimate to assume that the agent does not have any leverage on the behaviour of the world.

Things will become more clear if we replace the agent's actions with some events. It is legitimate to assume that the agent can choose his action, but it is not legitimate to assume that the agent can choose the event which is going to happen next.

Moreover, it is not perfectly legitimate to assume that the agent has full control on his actions. Consider the action "I graduate university" or "I play darts and hit ten". These actions are not entirely under our control. Therefore, it is more appropriate to articulate our preference rather than the exact event that will happen or the exact action we will do next.

What is *preference*? We have some constraint set by the model and within that constraint we have some unpredictable randomness. For example, the model tells us that the probability of rainfall must always be above 10% and below 80%. Thus, the model tells us that we can never reduce the rainfall probability below 10% or increase it above 80%. The model tell us the interval but not the exact probability in that interval. We are allowed to have some preference. We can prefer having some rain to water the crops or not having any rain so we can go to beach. Our preference can somehow influence the probability. We may take some action to drive the probability towards our preference, e.g. pray for rain or launch cloud-seeding missiles in the sky to trigger some rain.

Our preference for something to happen does not warrant that it will happen. Sometimes it can indeed be quite the opposite. The more we long for something to happen, the lesser the probability of our longings being granted. We will exert influence on the events through our preferences. The magnitude of that influence depends on the magnitude of the power we have. If our power is absolute then our preference will be "Royal". If the Royal preference is for rain, we will have rain with a probability of 80% (the maximum probability permitted by the model). If the Royal preference is for dry weather, we will have rain with a probability of 10% (the minimum probability).

The preference will not be part of the model. We have a model which tells us what can happen and some preference which indicates what we prefer to happen.

**Definition.** A preference for action is a function which for each state returns a list of preferred actions. The actions are listed in descending order from *most wanted* to *most unwanted*.

$$Preference : S \to List(\Sigma)$$

The preference for action will provide us with a deterministic policy (play the most preferred move). This will be the case when the action is not limited by any constraint. If there are constraints we will end up with a nondeterministic policy, which will be the Royal preference. E.g. if the constraints of the preference are *[a$_1$, b$_1$], [a$_2$, b$_2$], ...* then we play the first move with a probability of *b$_1$*, the second move with a probability of *(1-b$_1$).b$_2$* and so forth.

Similarly, we can define two other preferences: "preference for the response from the world" and "preference for the next event".

## MDP Plus Inversed

Can we inverse the MDP Plus model in order to obtain an MDP Plus model which predicts the past? The answer is yes, we can. We will again assume that there are no white peaks, otherwise the inverse MDP Plus model will not be unique.

We will use the same process that we used for inversing the FOMM. First, we will inverse MDP Fixed, in which the probability of each forward arrow is fixed. In the inversed model, the probability of each backward arrow will be fixed, too (if the model is without white peaks). For each state and inbound arrow entering that state there will be a fixed probability which tells us how likely is that we have entered the state exactly through that arrow. That probability must be split in two probabilities (one for the likelihood of that particular action and one for the likelihood of having exactly that arrow for that action). Such splitting is a straightforward process. This will produce an MDP Fixed model. Therefore, the inverse of MDP Fixed will again be an MDP Fixed model.

If we inverse an MDP Plus model, the probabilities of the arrows will be intervals rather than fixed values. How can we obtain arrow probability intervals in inverse direction? In an MDP Plus, the agent and the world have many possible policies. For each arrow, from all these policies we will select those which use the arrow most rarely. This will give us the minimum. Similarly, we will obtain the maximum by selecting the policies which use the arrow most often. Thus, we will obtain the interval of probabilities for each arrow. How can this interval be split in two intervals? Again, this is a straightforward process.

Interestingly, the inverse of an MDP model is not an MDP model (an inverse model exists, but it is MDP Plus). This is the reason why other authors do not consider an inverse MDP model which predicts the past. Let the agent enjoy absolute free will while the world is constrained to following a precisely defined policy (this is the MDP model). Then, inversing the arrows will give us an agent who is subject to some constraints and a world which enjoys a degree of freedom (i.e. the so-derived model is MDP Plus rather than MDP).

## The Markov property

Our next step will be to abandon the Markov property. This means we will dispense of the property that the model cannot be improved. The bottom-line is that we will discard perfection altogether and the so-derived model will be neither perfect nor quasi-perfect.

As we demonstrated for the FOMM, having the Markov property in place means that the model cannot be improved any further. This property implies that all facts which are relevant both for the past and for the future have been reflected (memorized). In other words, nothing else is worth to memorize. Everything worth memorizing has already been memorized. If there was such a fact we might improve the model by adding this fact to the already memorized ones (we will do that by increasing the number of states).

By abandoning the Markov property we move from a generator model to descriptor model. Now we do not say everything about the world, but only provide some statistical dependencies which partially describe the world. For example, knowing that "Monday" occurs with a probability of one-seventh is useful, but from this does not imply that the day after Sunday is Monday.

## The Event-Driven model

Now is time for the most important generalization. The agent's actions will be replaced with some events. This is the most prominent abandonment of perfection, because from now on we will not monitor the agent's actions at each step, but more monumental events some of which may occur quite rarely. The model will no more change its state at each step (with every action), but only upon the occurrence of an event monitored by the model.

The so-derived Event-Driven (ED) models will provide a very broad description of the world by saying no more than few things about the observed events. For example, an ED model can tell us whether we had dinner and then brushed our teeth or vice versa. This information is important, but in our world there are many other things which the ED model will forgo.

**Definition:** An Event-Driven model is:
$S$ (the set of states)
$E$ (the set of events monitored by the model)
$Trace : S \times \Omega \to \mathbb{R}$ (the probability of the agent seeing a concrete observation in a concrete state)
$Event : S \times E \to \mathbb{R}$ (the probability of a particular event to happen for one step)
$World : S \times E \times S \to \mathbb{R}$ (the probability of a transition from one state over a certain event to another state)

The trace here is still perfect. We will soon generalize it and make it imperfect.

For the sake of simplicity, in our definition the functions *Event* and *World* return probabilities rather than intervals. In fact, typically we will assume that the probability is not of any interest as we will only want to know if something can or cannot happen (i.e. in most cases the functions will return the interval [0, 1] or the values 1 or 0).

**Note:** Our definition of the Event-Driven model describes something which is almost perfect. Of course, we should add Markov property, because without it we cannot have perfection. If we are to have perfection, the events must be strictly defined, too. (E.g., they can be defined by characteristic functions which return only 0 or 1 and depend only on the past.) Furthermore, let us assume that two events cannot occur at the same time. We already assumed that the trace is perfect and that the probability returned by the functions *Event* and *World* is precisely defined. With these assumptions, the Event-Driven model will be perfect indeed and will be a generator which provides a perfect description of the world. Let us add that with this assumption for the *Event* function, either the events will not depend on the agent's actions or the agent will not have free will and will be bound to follow a fixed policy.

Our MDP model describes the world without saying anything about the agent (i.e. we let the agent loose so he can do whatever he likes). In MDP Fixed and MDP Plus we imposed some constrains on the agent's actions. Our ED model describes the world in its togetherness with the agent. The point is that we are part of the world and when exploring the world we also explore ourselves. If an event is impossible, the reason may be the world, but the reason may well be you (the agent). Something may be impossible to happen either because the world does not permit it or because you (the agent) do not want it to happen (or cannot make it happen even though you wish it to happen).

The nondeterminism of the Event-Driven model exceeds by far the nondeterminism of the MDP model. The agent's actions are events which do not intersect (cannot occur concurrently), while the events observed by the ED can perfectly occur at the same time. When this occurs, we need to decide which arrow we should follow. Thus, there is yet another source of nondeterminism. In addition to several arrows with the same colour there may be arrows with different colours, which leaves us wondering which arrow to follow since both events have occurred at the same moment of time. In the ED we can avoid such additional nondeterminism by assigning priorities to the events and thereby permit such collisions. Another solution in these situations is to use both arrows by going down one of the arrows and then going down the other one (this will acknowledge the occurrence of both events). Of course, in this case it must be decided which arrow will be used first.

We should not hope that in an ED model we will know which exactly is the current state of the model we are in. The nondeterminism of the model means that typically we will not know the exact answer of this question, but will know the answer with a margin of probability. Thus, we will often be asking the question "Where am I?" or "What is going on right now?" We can reduce these two questions to "In which state of the model I am now?".

## What is a trace?
A trace is what makes the various states distinct from each other. A model without a trace is futile. Imagine that everything is grey. So, will it matter which state are we in?

A trace has two functions. First, it makes meaning by telling us what is expected to happen and second it helps us to find were we are. (It helps us find out which the current state is. We may not know where we are because of nondeterminism.) This makes the trace an essential part of the definition of the model. Suffice it to change the trace only, keeping everything else unchanged, and we will end up with a very different model.

In the following example there is only one action and two possible observations (*red* and *blue*). Let us consider the model depicted in Figure 5.

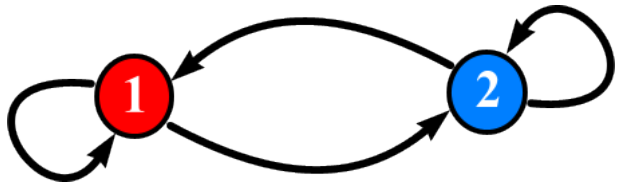

**Figure 5**

In this model we have not assigned probabilities to the arrows. We assume we do not know what the probability is, meaning that the probability is in the [0, 1] interval. Thus, the model is not MDP, but MDP Plus.

This model does not tell us anything about the world. It does not tell us even which the current state is, although we can figure it out from what we see at the moment: *red* or *blue*. More important than that is what will happen in the future or what has happened in the past. The model however keeps silent on these questions.

Nonetheless, the model is interesting. It is nondeterministic, but if we know what we are going to see at the next step we will be able to identify exactly the state we will be in and vice versa. I.e. there are two things we care about and we can derive the first one from the second or vice versa, but we have no way to learn either the first or the second. Thus, in the end of the day, the model is useless.

Let us now change the world so that our observations become {1, 2, 3, 4}. Instead of *red* or *blue*, let the trace be *even* or *odd*. (This trace is more general and goes beyond our earlier definition of the MDP Plus model, but we will shortly discuss even more complex trace versions.)

This gives us a new model which will be in state 1 *iff* the observation is even. Fine. Let us now change the trace again to *lesser than 3* or *greater than 2*. This gives us a very different model wherein the state is not determined by the odd or even number of the observation, but on whether the observation is *lesser* or *greater*.

## Imperfect trace

Now let us generalize the trace of the model and dispense of its perfection. A perfect trace is both total and complete. It is total because something is happening in each state. Complete means that we know what exactly is happening. After the generalization, the trace will not be total (present in each state) and will not tell us what exactly is going to happen.

A trace will be complete when we know what exactly we are going to see in the state. A trace is also complete when there are several possibilities and the trace gives us the exact probability of each possibility. The completeness can be disrupted by replacing the exactly defined probability with an interval (i.e. when we admit a degree of unpredictable randomness). The so-introduced intervals will cover the *even* trace we used above. In that case *even* will be 2 or 4, each one with a probability in the [0, 1] interval. Therefore, when the observation is *even*, we will not know whether it is 2 or 4, but what matters most is that it is not 1 or 3, because the probability of these is 0.

Once we replace the exact probabilities with intervals, we will be able to say almost everything about the trace which occurs at one step (at one moment of time). With MDP models, we stay in the state just one step. After that the next action occurs and we move to the next state. We created the Event-Driven models where we reside in the state not until the next step, but for a certain period of time (until the next monitored event occurs). Thus, instead of telling us what we can expect to see at a certain moment of time, the trace now tells us what might we see within a certain interval of time. An interval lets us see a lot more things than one point in time. (For example, we can see that the observation does not change and stays the same throughout the interval).

The most interesting statement we can make about an interval of time is that a certain phenomenon will be observed within that interval of time. Let is first say what is a phenomenon.

## Phenomenon

We introduced the Event-Driven models which are not perfect and do not tell us everything, but describe a certain dependency (pattern). The question is whether the pattern is observed all the time or only from time to time. Consider the pattern in Figure 6.

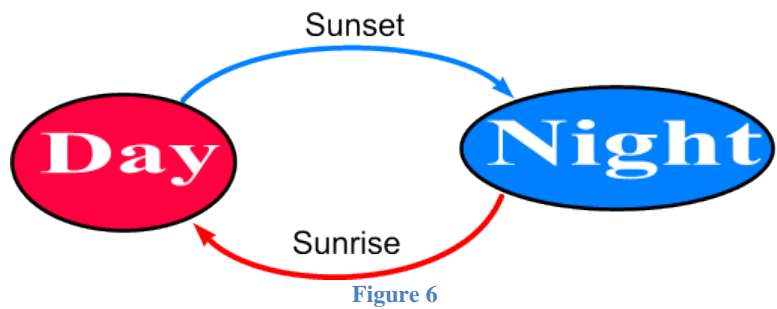

**Figure 6**

The pattern above will be observed all the time because it is always either day or night. Imagine now that we embark on a journey to planet Mars. Now the sun will shine all the time and we will break away from the day-night pattern. Hence, although some patterns are observed permanently, this is not the case for most other patterns.

**Definition.** A phenomenon is an Event-Driven model which is not valid permanently (throughout the entire life), but only during certain intervals of time.

Consider the phenomenon *rainstorm*. If we can describe it with an ED model, that model will not be valid all the time, but only when a storm is raging.

A concrete observation from $\Omega$ is also a phenomenon, because we can easily construct an ED model where this and only this phenomenon is observed all the time. (I.e. we will observe that observation *iff* that model is valid.)

The tests discussed in [4] can also be described by phenomena. The test consists of condition and result. The condition can be represented as a fact (set of states) and the result as the trace in those states. That is, we can describe the phenomenon: "The test would give a positive result."

Let us now define the opposite of a phenomenon:

**Definition**: A permanent pattern is an Event-Driven model which is permanently valid (throughout the entire life). In other words, the entire life of the agent can be expressed as a path in the ED model.

We will never know whether a permanent pattern will evolve into phenomenon. In other words, we will not know whether the validity of the event-driven model will come to an end. Furthermore, in the general case will not be sure whether at this moment we are indeed observing a given event because the fact that an event is being observed involves a degree of nondeterminism.

We can describe the trace of an Event-Driven model using certain phenomena (i.e. other ED models). The trace can be such that a phenomenon will be observed in a given state (or may be observed, or will be observed with a certain probability, etc.).

## Trace with a memory

As we said already, replacing the exact probabilities with intervals enables us say almost everything about the trace which occurs in one step. There are, however, certain things which we cannot tell. Now let the trace have memory and store in that memory the last observation in that state. Let the new observation in that state be the same as the previous one with a large degree of probability.

By way of example, let a house be the model of the world. Let the rooms in the house be the states of that model. The event will be "I move from one room to another". Let the lamps be switched on in some rooms and switched off in other rooms. The memory of the trace will store the status of the lamps: on or off. When we go back to a room in which the lamp was on before, probably the lamp will be on again unless somebody switched it off in the meantime.

Objects are also things which we should memorize. The emergence of an object is a phenomenon. It is perfectly natural that the object is present in some states and absent in others. Let us again have our house as a model of the world. Let's have in the house fixed objects, e.g. pieces of furniture such as sofas, and moving objects, e.g. people walking from one room to another. Thus, we need not memorize where the sofas are at the moment because they do not move around. Consequently, we can store the sofas in our fixed memory (in the trace of the model), while the people must be stored in our dynamic memory (the memory of the trace) so that we can remember where each person is at any moment of time. Thus, the memory of the trace will not be part of the model. There will be a fixed memory which stores the model and a dynamic memory which stores the current state of the model and the trace memory (in which room are we staying and where the moving objects are).

# Object

What is an object? In [6] we defined the object as an Event-Driven model. In other words, we associated the phenomenon "I am observing the object" with the object itself.

Here we will change the definition of an object. We will borrow from [3] which describes the object as an abstract thing characterized by certain properties.

**Definition.** A property is a phenomenon which occurs when we observe an object from the group of objects which possess that property.

In other words, the property will be an Event-Driven model, while the object will be an abstract notion which is characterized by some properties.

Can we say that if two objects have the same properties then these objects coincide? No, we cannot. Let's take a pair of twin brothers. This example is not very good because the two brothers are not exactly identical – at least they have different names. How about two identical kitchen chairs? Let the chairs be so identical that we cannot tell one from the other. Even so, they are two different objects.

# Reducing the number of states

Let us go back to our initial objective, namely to reduce the number of states in the model. We created the Event-Driven model which provides a rough description of the world. The ED does not say everything about the world and its states are much less than those of the generator model.

Can there be a relation between the generator and the ED model? Yes, we can express the ED model as the quotient set of some generator with respect to some equivalence relation.

How should this generator look like so that we can break it in classes of equivalence and thereby derive the ED model? Roughly said, the states in that generator should "know" enough. If the states of the Event-Driven model "know" something which the states of the generator do not, then there will be a state of the generator which should belong to two different classes of equivalence at the same time.

The exact requirement to the generator is: The arrows set of the generator must include all events monitored by the Event-Driven model. But this is not enough. Furthermore, the event "Moving from one equivalence class to another" must be covered by the monitored events (i.e. it must be a subset of the union of the monitored events). If that event is not covered, then we might move from one state to another even if none of the monitored events occurs.

That was a fairly complicated description of our requirements to the generator model. The good news however is that these requirements are not important because we will not attempt to find that model. We will proceed directly with our search for the Event-Driven model, but in the back of our mind we will know that any such model can be expressed as a quotient set of some generator model with respect to the equivalence relation "the two states correspond to the same state of the ED model".

**Theorem 2.** For each event $E$ there is an Event-Driven model which monitors that event.

**Proof:** We shall construct an Event-Driven model which has two states and describes the fact that event $E$ has occurred an even number of times. Then we will take a generator which includes event $E$ (such a generator does exist because this is a requirement of our definition of an *event*). Generally, we will not be able to break that model in equivalence classes separated by $E$, but we will construct another generator model.

Similar to the approach we used in proving Statement 1, we will construct an equivalent model by doubling the states of the first model. We will replace each state $s_i$ with $s_i'$ and $s_i''$. Each arrow $s_i \to s_j$ will be replaced with two arrows:

$s_i' \to s_j''$ and $s_i'' \to s_j'$, if $(s_i \to s_j) \in E$
$s_i' \to s_j'$ and $s_i'' \to s_j''$, if $(s_i \to s_j) \notin E$

The two equivalence classes will be the sets $S'$ and $S''$, where $S' = \{s \mid \exists i : s = s_i'\}$, same for $S''$.

∎

Certainly there is not just one, but many Event-Driven models which monitor event $E$.

Thus far we explained the relation between the generator and the Event-Driven model. Although we can derive the ED model as a quotient set of some generator, we will not take this road and instead we shall construct the ED model directly using real events.

## Real events

Before we can search for an Event-Driven model, we need to select several events which will be monitored in that model. We must somehow define these events and learn how to recognize them (i.e. detect them as soon as they occur).

We said that we will not go out looking for a generator and then break it down to classes of equivalence. Instead, we will try to find the Event-Driven models outright. To this end, we need to learn how to detect events. We will not use the theoretical definition of an *event* which is unfeasible to apply in practice. Instead, we will try to detect events using two approaches – direct and indirect.

For the direct approach we will use characteristic functions. For the indirect approach we will use the trace (i.e. from what is going on we will infer that have moved to a new state and thereby that an event has occurred).

The direct detection approach gives us the exact moment in which the event occurred and enables us describe models with loops (i.e. it may be that the occurrence of an event does not trigger a change of the state). With the indirect detection approach it is more difficult to establish the exact time of occurrence because the change of the trace may be detected either immediately or after a few steps. Furthermore, with indirect detection we cannot find loops because when the state remains the same, the trace also stays the same.

With the direct detection approach we select a characteristic function and go out looking for a trace. I.e., we assume we know when the event occurs and try to find periods of time before and after the occurrence which are somehow specific (i.e. something special should occur in these periods).

The indirect detection approach goes the other way around. First we look for specific periods. Then we associate these periods with various states of the Event-Driven model and look for events which occur at the boundary of these periods (i.e. we try to find characteristic functions). We may not necessarily find a characteristic function which describes the transition. It may that our event remains "invisible", i.e. one which can only be detected indirectly.

## The characteristic function

The classic characteristic function returns the values 0 and 1. However, we will assume here that it returns a probability because it would be rather restrictive to consider only characteristic functions which tell us exactly whether an event has or has not occurred. We prefer a function which is capable of saying that the event has occurred with a certain probability.

We will go even further and assume that the characteristic function returns a probability interval. For example, if the function says that the event has occurred with a probability greater than 1/2, it will return the interval [1/2, 1]. If the characteristic function cannot say anything, it will return the interval [0, 1].

Which will be the argument of the characteristic function? What will determine whether an event has or has not occurred? The arguments will be a possible development of the past (a development that has occurred) and a possible development of the future (a development that will occur).

Why have we chosen a characteristic function which is dependent on both the past and the future? Isn't it better that the function depends on the past only? Indeed, we prefer to know that an event has occurred as soon as that event occurs. In other words, we would like the characteristic function to depend on the past only, but oftentimes there are events about the occurrence of which we learn later (somewhere in the future). That is why our characteristic function is defined as dependent on both the past and the future.

What shall we do if the characteristic function returns different values for two different intervals of time? We shall assume that the value coming from the longer interval is more credible. We may even assume that as the interval of time becomes longer, the characteristic function becomes more precise (the probability interval shrinks). Nevertheless, will not make the latter assumption because sometimes having more information makes us less confident rather than more confident.

An example of a characteristic function is the function which describes one of the agent's actions. This function will only keep an eye on the next action of the agent and will return 1 (the agent did exactly this action) or 0 (the agent did something else). Consequently, the events of the MDP model can be described by characteristic functions.

## Conclusion

The gist of this article is that the world cannot be understood completely, so if wish to construct a model of the world we should forget about perfection and instead find simple models which describe the world partially.

Another assertion in the article is that a singular simple model cannot tell us everything about the world, which means that we should aim to find many different models, each one describing part of the world (a certain dependency, property or phenomenon). Certainly, even if we find many simple models, they will not tell us everything, either, but we hope they will tell us enough.

We introduced Event-Driven models. These are models which describe a tiny part of the world. An example of an ED model is provided in Figure 6. The only thing this model tells us is whether it is day or night. Although this information is important, it is grossly insufficient for understanding the world, because besides the day-or-night dilemma there are many other important things in the world.

As said already, we will use Event-Driven models to describe various dependencies, phenomena and properties. Then, on the basis of these phenomena and properties we will create abstractions such as objects and agents. As we said, we explain the free will of the world with the agents living in that world. For every human being in the world, we will consider that human to be both an object and an agent – an object which we can observe and an agent whose actions we can detect.

We said that rather than looking for a single perfect MDP model will aim to find a raft of neat simple Event-Driven models. How neat and simple? The number of events monitored by the model will typically be in the order of 1-2. The number of states in the ED will be in the order of ten or so. The question, then, is: How can we possibly describe a complex world with models that are so simple? The key to this is that we will construct the models hierarchically. From more simple models we will derive more complex models. As we said, the trace in an Event-Driven model can be characterized by phenomena, which in turn are other, more simple ED models. Furthermore, when a model comes to a certain state, that will be an event which can be monitored by another model (i.e. it can be used for the creation of a more complex Event-Driven model).

If we wish to find a model of the world, we should abandon the quest for a perfect model of the world. On a final note, we will recall Voltaire's aphorism: "Perfect is the enemy of good". This wisdom appears to hold true for Artificial Intelligence as well.

# References

[1] Xi-Ren Cao (2005). Basic Ideas for Event-Based Optimization of Markov Systems. *Discrete Event Dynamic Systems: Theory and Applications, 15, 169–197, 2005.*

[2] Xi-Ren Cao, Junyu Zhang (2008). Event-Based Optimization of Markov Systems. *IEEE TRANSACTIONS ON AUTOMATIC CONTROL, VOL. 53, NO. 4, MAY 2008.*

[3] Tatiana Kosovskaya (2019). Isomorphism of Predicate Formulas in Artificial Intelligence Problems. *International Journal "Information Theories and Applications", Vol. 26, Number 3, 2019, pp. 221-230.*

[4] Dimiter Dobrev (2017). How does the AI understand what's going on. *International Journal "Information Theories and Applications", Vol. 24, Number 4, 2017, pp.345-369.*

[5] Dimiter Dobrev (2019). Minimal and Maximal Models in Reinforcement Learning. *International Journal "Information Theories and Applications", Vol. 26, Number 3, 2019, pp. 268-284.*

[6] Dimiter Dobrev (2019). Event-Driven Models. *International Journal "Information Models and Analyses", Volume 8, Number 1, 2019, pp. 23-58.*